\documentclass[acmtog,nonacm]{acmart} 
\authorsaddresses{}

\AtBeginDocument{%
  }
\usepackage{booktabs}   
\usepackage{multirow}   
\usepackage{graphicx}   

\usepackage{xcolor}
\usepackage{colortbl}
\definecolor{tabhl}{HTML}{FCECE2}

\begin{document}


\title{NemoSplat: Feed-Forward 4D Gaussian Splatting for Media-Aware Underwater Reconstruction}
\author{Xiaopeng Guo}
\orcid{0009-0004-4982-5457}
\affiliation{%
  \institution{The Hong Kong University of Science and Technology}
  \city{Hong Kong}
  \country{China}
}
\email{xguoay@connect.ust.hk}

\author{Wai Chung Tse}
\affiliation{%
  \institution{The Hong Kong University of Science and Technology}
  \city{Hong Kong}
  \country{China}}
\email{larst@affiliation.org}

\author{Yipeng Zhu}
\affiliation{%
  \institution{The Hong Kong University of Science and Technology}
  \city{Hong Kong}
  \country{China}}
\email{cpalmer@prl.com}

\author{Hanwen Zhang}
\affiliation{%
  \institution{The Hong Kong University of Science and Technology}
  \city{Hong Kong}
  \country{China}}

\author{Huajian Huang}
\affiliation{%
  \institution{Beijing Institute of Technology}
  \city{Beijing}
  \country{China}}
\email{jsmith@affiliation.org}

\author{Sai-Kit Yeung}
\affiliation{%
  \institution{The Hong Kong University of Science and Technology}
  \city{Hong Kong}
  \country{China}}
\email{saikit@ust.hk}


\begin{abstract}

Reconstructing photorealistic scenes in unconstrained underwater environments remains challenging due to severe media-induced light scattering and unpredictable dynamic objects. Recent feed-forward visual foundation models have demonstrated remarkable capabilities in generalized novel view synthesis and tracking. However, when directly applied to aquatic videos, optical attenuation and motion interference fatally corrupt their feature aggregation, leading to severe tracking and reconstruction failures. To overcome these limitations, we present NemoSplat, the first feed-forward 4D Gaussian Splatting framework tailored for media-aware dynamic reconstruction directly from uncalibrated marine videos. Beyond providing robust estimations of camera poses and dense scene depth, we devise a Promptable Dynamic Disentangler that utilizes a confidence-aware fusion strategy of learned dynamic probabilities and optional semantic text priors, effectively isolating massive transient entities. Furthermore, to counteract visual degradation, a Media-Aware Gaussian Predictor is formulated to jointly estimate intrinsic 3D Gaussian attributes alongside physical media parameters, rendering pristine scene appearance in a single forward pass. Additionally, we introduce a large-scale underwater dataset with massive dynamic elements to facilitate training and evaluation. Extensive experiments on our dataset demonstrate that NemoSplat achieves state-of-the-art tracking accuracy and high-fidelity rendering.

\end{abstract}

\keywords{3D Gaussian Splatting, Feed-Forward Models, Underwater Media Modeling}
\begin{teaserfigure}
  \includegraphics[width=\textwidth]{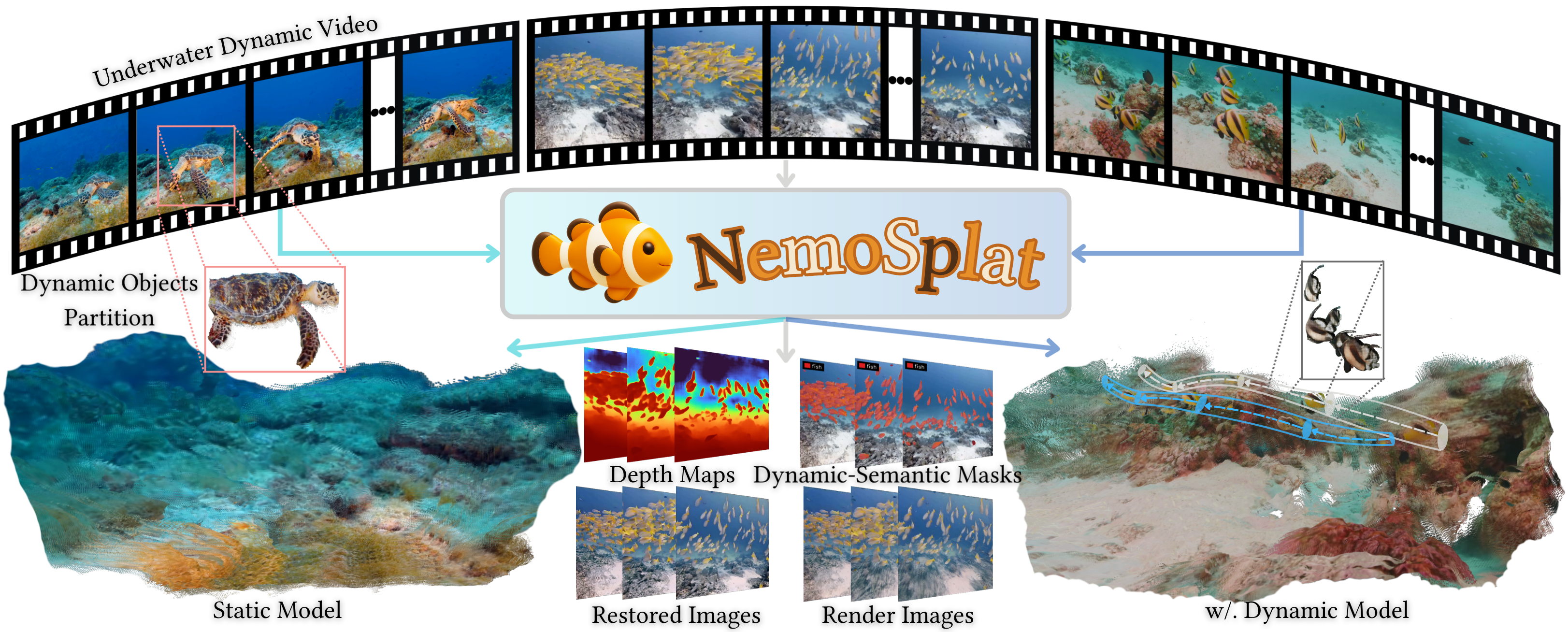}
  \Description{A teaser image showing the main result and concept of the paper.}
  \caption{ 
  We present NemoSplat, a novel feed-forward model that directly reconstructs photorealistic underwater scenes from uncalibrated image sequences, robustly handling numerous dynamic objects and water-induced degradation.
  }
  \label{fig:teaser}
\end{teaserfigure}


\maketitle

\section{Introduction}


Robust visual pose estimation and photorealistic reconstruction of underwater scenes are essential for marine applications such as biological monitoring, autonomous navigation, and ecological conservation~\cite{underwater_survey}. However, authentic underwater environments pose formidable challenges for conventional geometric and vision-based reconstruction pipelines. The inherently dynamic ecosystems teeming with moving entities and severe media-induced visual degradations, such as light absorption and scattering~\cite{seathru}, make unconstrained underwater reconstruction notoriously difficult.


Recently, the emergence of feed-forward visual foundation models~\cite{dust3r,mast3r,vggt,pi3,streamvggt} has enabled robust geometry and pose estimation from uncalibrated videos. Building upon these geometric foundations, feed-forward 3D Gaussian Splatting (3DGS) approaches~\cite{mvsplat, pixelsplat, anysplat} extend point cloud reconstruction to photorealistic rendering in a single forward pass without per-scene optimization. However, being intrinsically designed for static, clear-air environments, when directly deployed in marine domains, their feature aggregation mechanisms are fatally corrupted by underwater motion interference and optical attenuation, leading to tracking failures and reconstruction degradation. Consequently, unlocking the potential of feed-forward models for marine applications requires a novel paradigm that explicitly disentangles transient dynamics from media-induced degradation.

Driven by these critical limitations, our core motivation is to architect a generalized, pose-free feed-forward framework capable of simultaneously tackling complex aquatic dynamics and severe media-induced visual degradation. As depicted in Fig.~\ref{fig:teaser}, we present NemoSplat, the first feed-forward 4DGS visual foundation model tailored for media-aware dynamic reconstruction directly from uncalibrated marine videos. At its structural core, the elegance of NemoSplat is propelled by meticulously designed mechanisms that holistically resolve both geometric motion and optical scattering. As a foundational prerequisite, our network reliably extracts accurate camera tracking trajectories and dense scene depth to geometrically anchor the uncalibrated visual inputs. Building upon this stable geometric baseline, we devise a Promptable Dynamic Disentangler that employs a confidence-aware logit fusion to fuse learned dynamic probabilities with optional semantic priors, inextricably decoupling precise dynamic masks from the static environment. Concurrently, we utilize a Media-Aware GS Predictor to jointly estimate intrinsic 3D Gaussian attributes alongside physical media parameters. By yielding all essential representations in a single forward pass, our framework implements a principled compositional strategy, utilizing the static geometry as a structural base while superimposing time-varying dynamic primitives to simultaneously reconstruct the full 4DGS scene and recover the pristine underwater imagery.

Furthermore, to facilitate robust training and rigorous evaluation in this under-explored domain, we curate a comprehensive dataset of dynamic underwater sequences. Encompassing diverse marine scenes with varying degrees of optical attenuation and complex object movements, this dataset provides an essential benchmark for advancing research in aquatic dynamic scene reconstruction. 

To summarize, our main contributions include:
\begin{itemize}
    \item We introduce NemoSplat, the first feed-forward 4DGS framework for media-aware reconstruction in dynamic underwater environments, achieving state-of-the-art performance in geometry estimation and novel view synthesis.
    
    
    

    

    \item We propose Promptable Dynamic Disentangler, fusing dynamic probability with optional text semantic prior to enable the robust disentanglement of massive moving objects from static backgrounds.
    
    \item We formulate Media-Aware Gaussian Predictor, a physics-embedded rendering network regulated by multi-view consistency and physical losses, to inherently model water media and recover pristine appearances from degraded inputs.
    
    \item We construct a large-scale, dynamic underwater dataset with to facilitating underwater visual reconstruction method training and evaluation.  
    
\end{itemize}

\section{Related Work}\label{sec:relatedwork}

\subsection{Visual SLAM and Feed-Forward Reconstruction}
Visual SLAM systems have evolved from tracking handcrafted constraints~\cite{orbslam,dso} to leveraging robust deep visual priors~\cite{droidslam, mast3rslam}. Recently, visual geometry foundation models~\cite{dust3r, mast3r, vggt, streamvggt} unified the extraction of camera poses and scene depth via robust attention mechanisms, bypassing traditional pre-calibration needs, while their outputs remaining restricted to discrete geometric spaces lacking photorealistic synthesis capability. To recover high-fidelity scenes, modern SLAM pipelines~\cite{photoslam, monogs, splatam} integrate 3D Gaussian Splatting (3DGS)~\cite{3DGS}, but their mapping backends fundamentally necessitate continuous, iterative online optimization. Pioneering feed-forward 3DGS approaches, such as PixelSplat~\cite{pixelsplat} and AnySplat~\cite{anysplat}, bridge this gap by directly regressing 3DGS parameters in a single pass. Nevertheless, they fundamentally rely on static, clear-air assumptions. When confronted with severely degraded, dynamic underwater environments, their tracking resilience and feature matching pipelines inevitably collapse. 

\begin{figure*}[t]
  \centering
  \includegraphics[width=0.95\linewidth]{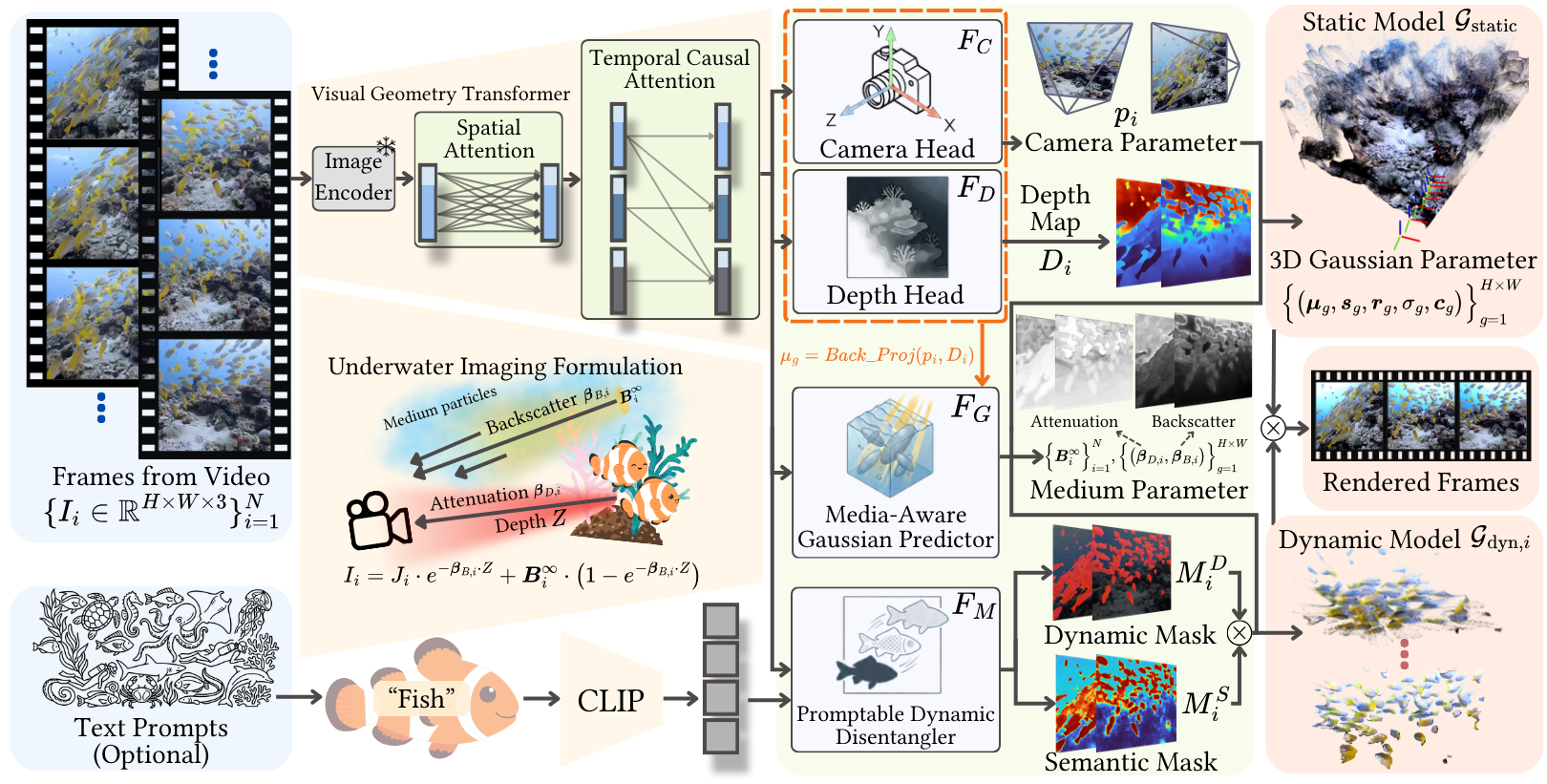}
  \Description{Overview figure.}
  \caption{Overview of our NemoSplat framework. It leverages the Visual Geometry Transformer to extract robust spatiotemporal features from uncalibrated underwater sequences. Subsequently, a shared feature encoder feeds these representations into specialized decoders to jointly estimate camera poses, depth, dynamic semantic masks (optionally refined via grounded text prompts), water backscatter and attenuation parameters, and intrinsic  4D Gaussians.
  }
  \label{fig:pipe}
\end{figure*}

\subsection{Dynamic Scene Modeling} 
Some dynamic novel view synthesis approaches~\cite{dynamic_3dgs, 4dgs} typically estimate time-conditioned deformation fields or rely on explicit optical flow to handle moving objects. In parallel, optimization-based dynamic GS-SLAM systems~\cite{DROIDW, WildGS-SLAM} mitigate the interference of dynamic objects by uncertainty-based down-weighting. However, these methods still rely heavily on camera priors, which must be estimated via Structure-from-Motion (SfM) modeling in uncalibrated underwater scenes. While recent 4D feed-forward networks~\cite{easi3r, more, 4dgt} use spatiotemporal attention to implicitly separate dynamic objects or introduce 4D Gaussian primitives, they struggle with the numerous small, cluttered distractors typical of underwater scenes, leading to motion leakage and ghosting. In contrast, our approach leverages high-level text-guided semantics to explicitly decouple dynamic masks, bypassing fragile low-level cues to secure a clean static background for accurate 4D modeling.

\subsection{Media-Aware Underwater Reconstruction} 
To address color degradation in scattering media, some physics-based rendering methods incorporate optical propagation models~\cite{seathru} into neural representations~\cite{seathru-nerf, seasplat, watersplatting}. These methods model light absorption and backscatter to recover clear appearance, but their inverse problem remains strongly ill-posed and typically requires expensive test-time optimization with external depth priors. More critically, they are largely formulated under static-scene assumptions. In unconstrained marine videos, dense dynamic agents violate this assumption and entangle transient motion with medium scattering, often causing geometric instability and erroneous color restoration. Here, our NemoSplat introduces a novel feed-forward media-aware 4DGS framework that jointly resolves both scattering and dynamics without test-time optimization. 

\section{Method}

We propose NemoSplat, the first feed-forward 4DGS framework tailored for rapid and high-fidelity scene reconstruction in unconstrained underwater environments. As illustrated in Fig.~\ref{fig:pipe}, given uncalibrated marine videos, our model integrates geometry prediction, dynamic-static separation and physical media restoration into a single forward pass, estimating robust camera tracking, precise dynamic masks, Gaussian primitives, and physical media parameters concurrently to synthesize pristine novel views.

In the following sections, we first define our mathematically decoupled problem formulation and the explicit underwater imaging model in Sec.~\ref{sec:problem}. We then detail the three core pillars of our architecture alongside their respective joint optimization objectives: the Geometry Estimator (Sec.~\ref{sec:geometry}), the Promptable Dynamic Disentangler (Sec.~\ref{sec:dynamic_disentangler}), and the Media-Aware Gaussian Predictor (Sec.~\ref{sec:media_gs}).

\subsection{Problem Formulation}
\label{sec:problem}

Given uncalibrated underwater images $\{I_i \in \mathbb{R}^{H \times W \times 3}\}_{i=1}^N$ and an optional text prompt $T$, our goal is to learn a direct neural mapping $f_{\boldsymbol{\theta}}$ that jointly infers camera tracking and scene representation without per-scene optimization. We systematically structure the network's outputs into per-frame and per-pixel levels, defining the global neural mapping $f_{\theta}$ as:
\begingroup\scriptsize
\begin{equation}
    f_{\theta}: \; \left(\{I_i\}_{i=1}^N, T\right) \;\longmapsto\; \Bigl\{ \big(p_i, D_i, M_i, \boldsymbol{B}_i^{\infty}\big) \Bigr\}_{i=1}^N \;\cup\; \Bigl\{ \big(\boldsymbol{\mu}_g, \boldsymbol{s}_g, \boldsymbol{r}_g, \sigma_g, \boldsymbol{c}_g\big), \big(\boldsymbol{\beta}^D_g, \boldsymbol{\beta}^B_g \big) \Bigr\}_{g=1}^{H\times W}.
\end{equation}
\endgroup

For each frame $I_i$, the model extracts the per-frame camera pose and environment state: $(p_i, D_i, M_i, \boldsymbol{B}_i^{\infty})$, where $p_i \in \mathbb{R}^9$ denotes the camera parameters (intrinsic and extrinsic), $D_i \in \mathbb{R}^{H \times W}$ is the depth map, $M_i \in [0,1]^{H \times W}$ represents the semantically-refined dynamic mask selectively guided by the input text prompt $T$, and $\boldsymbol{B}_i^{\infty} \in \mathbb{R}^3$ describes the global background veiling light. At the per-pixel level, the network decodes pixel-wise composite 3D Gaussian primitives. Each primitive $g$ explicitly bounds standard 3DGS~\cite{3DGS} geometric and photometric attributes: 3D center $\boldsymbol{\mu}_g \in \mathbb{R}^3$, scaling factors $\boldsymbol{s}_g \in \mathbb{R}^3$, rotation quaternion $\boldsymbol{r}_g \in \mathbb{R}^4$, opacity $\sigma_g \in [0,1]$, and view-dependent color $\boldsymbol{c}_g \in \mathbb{R}^{3 \times (K+1)^2}$ parameterized by degree-$K$ spherical harmonics. Specifically, $\boldsymbol{\mu}_g$ is deterministically obtained by back-projecting the 2D pixel to the world space using the estimated camera pose $p_i$ and reference depth $D_i$.
Our primitive formulation embeds two additional water medium parameters: the attenuation coefficient $\boldsymbol{\beta}^D_g \in \mathbb{R}^3$ and the backscatter coefficient $\boldsymbol{\beta}^B_g \in \mathbb{R}^3$.

Conditioned on the dynamic mask $M_i$, we formulate a structurally decoupled paradigm to explicitly isolate transient motion from the static scene. Let $\mathcal{G}_{\text{all}}$ encompass the complete set of Gaussian primitives predicted across the sequence. We systematically partition these into a global static set $\mathcal{G}_{\text{static}} = \{ g \in \mathcal{G}_{\text{all}} \mid M_i(g) < \tau \}$ and a collection of dynamic per-frame sets $\{ \mathcal{G}_{\text{dyn}, i} \}_{i=1}^N$, where each $\mathcal{G}_{\text{dyn}, i} = \{ g \in \mathcal{G}_i \mid M_i(g) \ge \tau \}$ using a gating threshold $\tau$. By rasterizing from the collective union $\mathcal{G}_{\text{static}} \cup \mathcal{G}_{\text{dyn}, i}$, we formulate the clean pristine radiance $J_{GS, i}$ for the $i$-th target frame.
Utilizing independent physical parameters~\cite{seathru}, we compose $J_{GS, i}$ and reconstruct the final observed color $\hat{I}_{\text{obs}, i}$, effectively striving to recover the pristine water-free image from the degraded inputs:
\begin{equation}
    \hat{I}_{\text{obs}, i} = J_{GS, i} \odot e^{-\boldsymbol{\beta}^D_i D_{i}} + \boldsymbol{B}_i^{\infty} \odot (1 - e^{-\boldsymbol{\beta}^B_i D_{i}}).
\label{eq:underwater_formation}
\end{equation}


\subsection{Geometry Estimator}
\label{sec:geometry}

To instantiate the global mapping $f_{\boldsymbol{\theta}}$ end-to-end (Fig.~\ref{fig:pipe}), our geometry estimator employs a Streaming Geometry Transformer for temporal encoding, coupled with dedicated decoder heads for camera pose and dense depth regression.

\paragraph{Streaming Geometry Transformer.}
To process marine videos efficiently, we adopt a streaming transformer architecture~\cite{streamvggt}. Each frame is tokenized via DINOv2~\cite{dinov2} into image tokens with $14\times 14$ size, a learnable camera token, and 4 standard register tokens. We replace memory-intensive global temporal attention~\cite{vggt} with an $L$-layer transformer alternating between spatial intra-frame attention and causal cross-frame attention. By querying a constant-memory key-value (KV) cache of past frames, the cross-frame module efficiently aggregates multi-view geometry priors. This streaming design scales seamlessly to long marine sequences, yielding temporally coherent token representations for downstream regression.

\paragraph{Camera Pose Head.} 
Following the visual backbone, 9-DoF camera parameters $p_i$ are regressed from dedicated camera register tokens. A standalone camera head, denoted as $F_C$, composed of a Multi-Layer Perceptron, decodes the camera tokens at the final transformer layer, predicting both camera extrinsics and intrinsics. The coordinate system is anchored to the first view (set as the identity transformation), with all other poses represented within this shared space. During optimization, we employ a pre-trained visual geometry transformer~\cite{streamvggt} as a teacher network, imposing Huber losses on the 9-DoF camera poses:
\begin{equation}
    \mathcal{L}_{\text{pt}} = \text{Huber}(p, p_{\text{teacher}}).
\end{equation}

\paragraph{Depth Head.} 
Dense scene geometry is decoded from the spatial patch tokens using a Dense Prediction Transformer (DPT) architecture~\cite{dpt}. This dedicated depth head ($F_D$) projects the tokenized spatiotemporal features into an original-resolution reference depth map $\hat{D}_{\text{dpt}}$, which serves as the fundamental geometric anchor for unprojecting and localizing the subsequent 3D Gaussian primitives in 3D space. To train this depth predictor robustly without ground-truth scans, we constrain $\hat{D}_{\text{dpt}}$ using confidence-weighted depths ($\mathbf{C}_{\text{teacher}}$) from the teacher network~\cite{streamvggt} alongside a Scale-Shift Invariant (SSI) structure loss supervised by spatial pseudo-labels from Depth-Anything-3~\cite{depthanything3} ($D_{\text{DA3}}$), where $\Delta D = \log \hat{D}_{\text{dpt}} - \log D_{\text{DA3}}$:
\begin{align}
    \mathcal{L}_{\text{dt}} &= \mathbf{C}_{\text{teacher}} \odot \text{Huber}(\hat{D}_{\text{dpt}}, D_{\text{teacher}}), \\
    \mathcal{L}_{\text{ssi}} &= \frac{1}{HW} \sum (\Delta D)^2 - \frac{1}{(HW)^2} \left( \sum \Delta D \right)^2.
\end{align}
\normalsize

The complete geometry objective $\mathcal{L}_{\text{geo}}$ orchestrating the pose and depth networks is therefore summarized as:
\begin{equation}
    \mathcal{L}_{\text{geo}} = \lambda_{\text{pt}}\mathcal{L}_{\text{pt}} + \lambda_{\text{dt}}\mathcal{L}_{\text{dt}} + \lambda_{\text{ssi}}\mathcal{L}_{\text{ssi}}.
\end{equation}
\normalsize

\begin{figure}[!t]
  \centering
  \includegraphics[width=\linewidth]{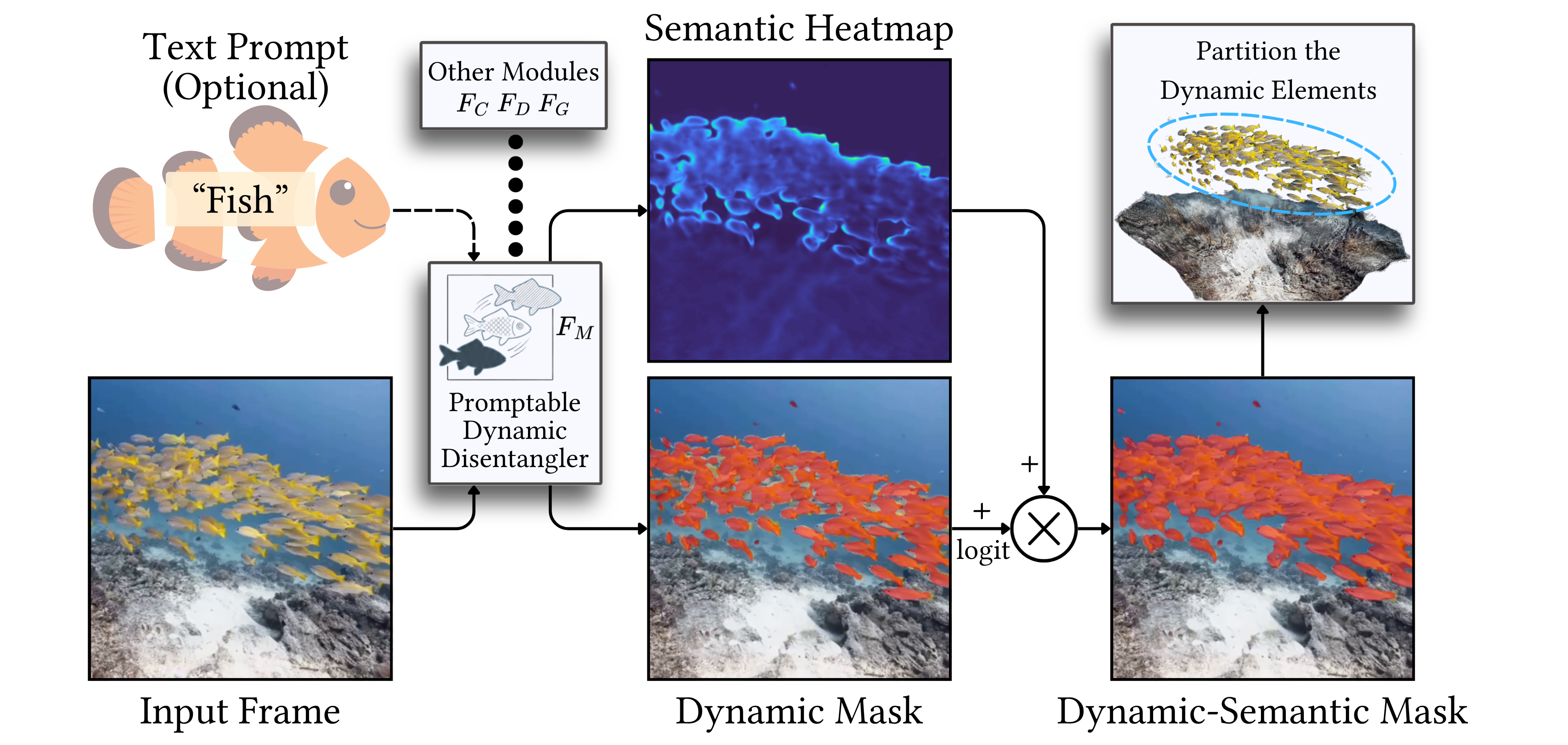}
  \Description{A technical flowchart showing the Dynamic-Semantic Mask Generation pipeline.}
  \caption{Dynamic-Semantic Mask Generation. The Promptable Dynamic Disentangler ($F_M$) refines the predicted dynamic mask via logit adjustment using an optional text prompt-guided semantic map, effectively revising a highly accurate mask for downstream 4D modeling.}
  \label{fig:mask_generation}
\end{figure}

\subsection{Promptable Dynamic Disentangler}
\label{sec:dynamic_disentangler}
 

Facilitating the explicit separation of dynamic entities from the background, we propose a versatile Promptable Dynamic Disentangler $F_M$. Leveraging a DPT-based framework, $F_M$ projects spatial patch tokens from the transformer backbone into a dense, full-resolution feature map. In parallel, a shallow three-layer $3\times 3$ CNN extracts high-frequency appearance cues directly from the raw image $I_i$, which are fused into the DPT features via element-wise residual addition, supplementing the fine edge details.

The merged representation is then passed through a convolution network that jointly predicts an initial dynamic object probability map $\hat{M}_{\text{dyn}}$ and a background water mask $\hat{M}_{\text{wat}}$. Crucially, this predicted water mask is subsequently forwarded to the downstream Media-Aware GS Predictor (Sec.~\ref{sec:media_gs}) to spatially constrain the learning of water media parameters. These structural priors are supervised using high-quality pseudo-labels $M^*$ generated by the Segment Anything Model 3~\cite{sam3}. To ensure sharp boundary discernment, we optimize both masks using a linear combination of Binary Cross-Entropy (BCE) and Dice loss~\cite{dice}. For the water mask, we explicitly apply a spatial boundary-proximity weight $\mathbf{W}_{\text{bound}}$ to penalize predictions near image edges:
\begin{align}
    \mathcal{L}_{\text{dyn}} &= \text{BCE}(\hat{M}_{\text{dyn}}, M^*_{\text{dyn}}) + 0.5 \, \text{Dice}(\hat{M}_{\text{dyn}}, M^*_{\text{dyn}}), \\
    \mathcal{L}_{\text{wat}} &= \mathbf{W}_{\text{bound}} \odot \text{BCE}(\hat{M}_{\text{wat}}, M^*_{\text{wat}}) + 0.5 \, \text{Dice}(\hat{M}_{\text{wat}}, M^*_{\text{wat}}).
\end{align}
\normalsize

\begin{figure*}[t]
  \centering
  \includegraphics[width=\linewidth]{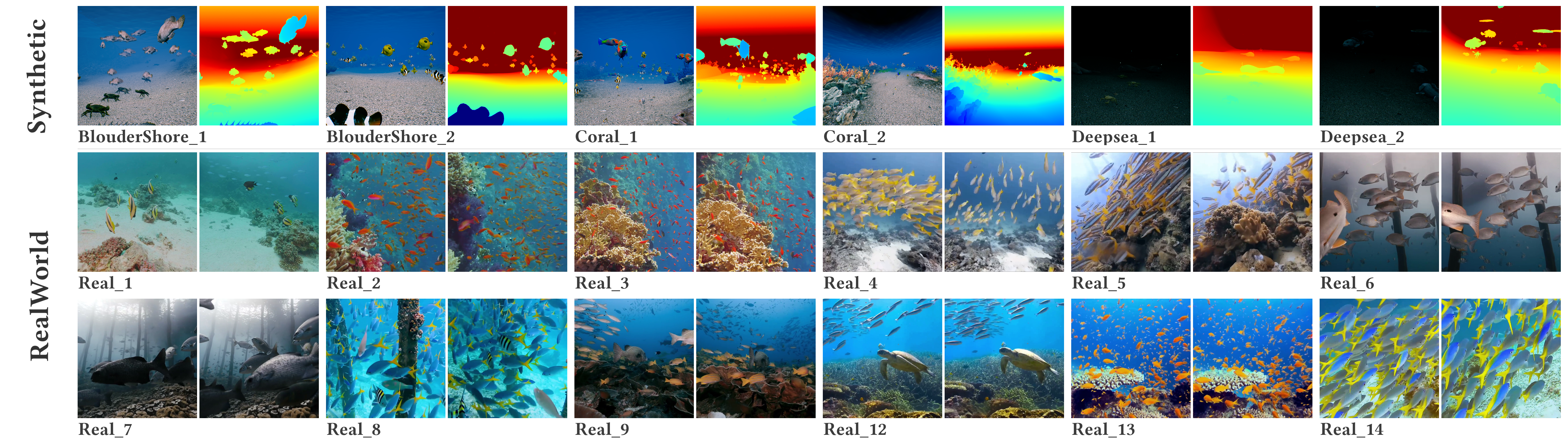}
  \Description{Sampled frames and depth maps from the evaluation dataset.}
  \caption{Examples sequences of our proposed evaluation dataset. It includes synthetic scenes with precise ground-truth annotations and diverse in-the-wild underwater sequences for comprehensive evaluation.}
  \label{fig:eval-set}
\end{figure*}

Our disentangler optionally accepts an arbitrary text prompt $T$ (e.g., ``fish'') to leverage explicit user intent, thereby extracting a highly refined and accurate final dynamic mask $M_i$. When such a prompt is provided,
a sibling semantic branch with the same convolutional structure projects the merged representation into a
$D$-dimensional embedding map $\hat{\mathbf{E}} \in \mathbb{R}^{H\times W\times D}$, which is $\ell_2$-normalized along the channel axis. Dense cosine similarities against the pre-trained CLIP~\cite{clip} text embedding $\mathbf{E}_{\text{CLIP}}$
(also $\ell_2$-normalized) then reduce to a single dot-product per pixel,
yielding a continuous semantic guidance map $S_{\text{sem}}$. During training,
this branch is supervised by the cosine-similarity loss:
\small
\begin{equation}
    \mathcal{L}_{\text{sem}} = 1 - \hat{\mathbf{E}} \cdot \mathbf{E}_{\text{CLIP}}.
\end{equation}
\normalsize

During inference, if a text prompt is supplied, we propose a strategy to refine the initial dynamic predictions by synergizing the base continuous geometric probability $P_{\text{dyn}}$ (derived from $\hat{M}_{\text{dyn}}$) with the semantic prior $P_{\text{sem}}$ (derived from $S_{\text{sem}}$), as Fig.~\ref{fig:mask_generation} shows. To address the inherent limitations of standard boolean logic, where strict intersections risk structurally destroying cohesive objects due to semantic noise, we avoid naively multiplying probabilities. Instead, we apply a confidence-aware adjustment mechanism to construct the final refined probability $P_{\text{final}}$:
\begin{equation}
    \text{logit}(P_{\text{final}}) = \text{logit}(P_{\text{dyn}}) + \gamma (P_{\text{sem}} - \tau_s),
\end{equation}
\normalsize
where $\gamma$ scales the semantic influence, and $\tau_s$ acts as a threshold. Specifically, the geometric log-odds are dynamically modulated by the semantic penalty term. Semantic scores above the threshold $\tau_s$ reinforce dynamic classification, while those below apply a suppressive penalty. This bidirectional logit adjustment efficiently dampens unprompted distractors while preserving boundary cohesion for accurate mask disentanglement. Notably, without explicit text prompts, the network seamlessly defaults to the geometry-driven $P_{\text{dyn}}$, functioning as a fully automatic dynamic-static separator.

\subsection{Media-Aware Gaussian Predictor}
\label{sec:media_gs}

Consistent with our pixel-wise dense strategy, we design a unified Media-Aware Gaussian Predictor (denoted as $F_G$) to jointly estimate the intrinsic 3D Gaussian attributes alongside the complex physical underwater properties in a single forward pass. 

Following the Gaussian head design of AnySplat~\cite{anysplat}, for each primitive unprojected from the reference depth map $\hat{D}_{\text{dpt}}$, this network densely regresses complete set of standard 3DGS~\cite{3DGS} parameters ($\sigma_g$, $\boldsymbol{s}_g$, $\boldsymbol{r}_g$, $\boldsymbol{c}_g$). Concurrently, to explicitly model volumetric scattering, the network maps the spatial features into the governing parameters of the underwater image formation model (Eq.~\ref{eq:underwater_formation}): defining the per-pixel direct signal attenuation $\boldsymbol{\beta}^D_i$, volumetric backscatter $\boldsymbol{\beta}^B_i$, and global per-frame veiling light $\boldsymbol{B}_i^{\infty}$.

Crucially, instead of predicting appearance and medium properties in isolated steps, we formulate a holistic media-aware training loss. Building upon the structure-decoupled formulation defined in Sec.~\ref{sec:problem}, we utilize the clean pristine radiance $J_{GS, i}$ rasterized exclusively from the collective union $\mathcal{G}_{\text{static}} \cup \mathcal{G}_{\text{dyn}, i}$. We then mathematically compose $J_{GS, i}$ with the predicted underwater parameters ($\boldsymbol{\beta}^{D}, \boldsymbol{\beta}^{B}, \boldsymbol{B}^{\infty}$) via the underwater image formation model (Eq.~\ref{eq:underwater_formation}) to calculate a joint rendering loss against the raw views:
\small
\begin{equation}
    \mathcal{L}_{\text{phys-render}} = \sum_i \left( \| I_{\text{raw}, i} - \hat{I}_{\text{obs}, i} \|_2^2 + \lambda_{\text{lpips}} \text{LPIPS}(I_{\text{raw}, i}, \hat{I}_{\text{obs}, i}) \right).
\end{equation}
\normalsize

To rigorously prevent the network from incorrectly baking water colors directly into the isolated Gaussian textures $J_{GS, i}$, we further regularize the medium properties. Inspired by prior physical optimization constraints in SeaSplat~\cite{seasplat} and WaterSplatting~\cite{watersplatting}, on distant water mask regions $\Omega_{\text{far}}$, we enforce $\boldsymbol{B}^{\infty}$ explicitly:
\small
\begin{align}
    \mathcal{L}_{B^{\infty}} &= \left\| \boldsymbol{B}^{\infty} - \frac{1}{|\Omega_{\text{far}}|} \sum_{u \in \Omega_{\text{far}}} I_{\text{raw}}^{u} \right\|_1, \\
    \mathcal{L}_{J} &= \left\| \frac{I_{\text{raw}} - \boldsymbol{B}^{\infty} \odot (1 - e^{-\boldsymbol{\beta}^{B} \hat{D}_{\text{dpt}}})}{e^{-\boldsymbol{\beta}^{D} \hat{D}_{\text{dpt}}}} - J_{GS} \right\|_1.
\end{align}
\normalsize
Furthermore, to respect strict physical marine optics, we follow their practices and integrate oceanographic Jerlov penalties regulating color channel attenuations ($\beta^D_{R} > \beta^D_{G} > \beta^D_{B}$) alongside a spatial Total Variation ($\mathcal{L}_{\text{TV}}$) on the parameters $\boldsymbol{\beta}$:
\begin{align}
    \mathcal{L}_{\text{Jerlov}} &= \max(0, \beta^D_{G} - \beta^D_{R}) + \max(0, \beta^D_{B} - \beta^D_{G}), \\
    \mathcal{L}_{\text{TV}} &= \|\nabla_x \boldsymbol{\beta}\|_1 + \|\nabla_y \boldsymbol{\beta}\|_1.
\end{align}
\normalsize

We explicitly enforce multi-view geometric consistency, ensuring that the rendered 3DGS topology successfully recovers the image structure. Specifically, we constrain the rasterized 3DGS depths ($\hat{D}_{\text{gs}}$) using the structure-aware DPT-predicted depths via an $L_1$ penalty:

\begin{equation}
    \mathcal{L}_{\text{dc}} = \| \hat{D}_{\text{dpt}} - \hat{D}_{\text{gs}} \|_1.
\end{equation}
\normalsize
Ultimately, the rendering optimization is harmonized as a comprehensive objective encompassing visual appearance, structural geometry, and explicit physical alignment:
\begin{align}
    \mathcal{L}_{\text{render}} &= \lambda_{\text{phys}}\mathcal{L}_{\text{phys-render}} + \lambda_{\text{dc}}\mathcal{L}_{\text{dc}} \nonumber \\
    &\quad + \lambda_{\text{water}} (\lambda_{B^{\infty}}\mathcal{L}_{B^{\infty}} + \lambda_{J}\mathcal{L}_{J} + \lambda_{\text{Jerlov}}\mathcal{L}_{\text{Jerlov}} + \lambda_{\text{TV}}\mathcal{L}_{\text{TV}}).
\end{align}

\normalsize

\begin{table*}[t]
\centering
\caption{Quantitative evaluation on synthetic sequences. Best results are \colorbox{tabhl}{\textbf{bold}}, second best are \underline{underlined}. ``-'' indicates that the method does not support rendering, and ``\(\times\)'' denotes failure due to out of memory. ATE is reported in meters.}
\label{tab:exp_synthetic}
\renewcommand\arraystretch{1.2}
\setlength{\tabcolsep}{2.5pt} 
\resizebox{0.90\textwidth}{!}{
\begin{tabular}{lcccccccccccccccc}
\hline
\textbf{Method} & \multicolumn{4}{c}{\textbf{BoulderShore}} & \multicolumn{4}{c}{\textbf{Coral}} & \multicolumn{4}{c}{\textbf{Deepsea}} & \multicolumn{4}{c}{\textbf{Average}} \\ 
\cmidrule(r){1-1} \cmidrule(lr){2-5} \cmidrule(lr){6-9} \cmidrule(lr){10-13} \cmidrule(l){14-17}
Metrics & ATE\(\downarrow\) & PSNR\(\uparrow\) & SSIM\(\uparrow\) & LPIPS\(\downarrow\) & ATE\(\downarrow\) & PSNR\(\uparrow\) & SSIM\(\uparrow\) & LPIPS\(\downarrow\) & ATE\(\downarrow\) & PSNR\(\uparrow\) & SSIM\(\uparrow\) & LPIPS\(\downarrow\) & ATE\(\downarrow\) & PSNR\(\uparrow\) & SSIM\(\uparrow\) & LPIPS\(\downarrow\) \\
\cmidrule(r){1-1} \cmidrule(lr){2-5} \cmidrule(lr){6-9} \cmidrule(lr){10-13} \cmidrule(l){14-17}
VGGT~\cite{vggt}        & 0.19 & - & - & - & \cellcolor{tabhl}\textbf{0.23} & - & - & - & 5.93 & - & - & - & 2.12 & - & - & - \\
StreamVGGT~\cite{streamvggt}  & 0.40 & - & - & - & 0.79 & - & - & - & 5.01 & - & - & - & 2.06 & - & - & - \\
WildGS-SLAM~\cite{WildGS-SLAM} & \underline{0.12} & 18.80 & \underline{0.60} & 0.45 & 4.97 & 17.01 & 0.55 & 0.52 & \underline{0.81} & \underline{29.86} & \underline{0.81} & \underline{0.35} & 1.96 & 21.89 & \underline{0.65} & 0.44 \\
Droid-W~\cite{DROIDW}     & \cellcolor{tabhl}\textbf{0.11} & 20.04 & \cellcolor{tabhl}\textbf{0.71} & 0.44 & 4.65 & 18.56 & \cellcolor{tabhl}\textbf{0.58} & 0.56 & \cellcolor{tabhl}\textbf{0.78} & \cellcolor{tabhl}\textbf{31.93} & \cellcolor{tabhl}\textbf{0.83} & \cellcolor{tabhl}\textbf{0.30} & \cellcolor{tabhl}\textbf{1.85} & \underline{23.51} & \cellcolor{tabhl}\textbf{0.71} & 0.43 \\
YoNoSplat~\cite{yonosplat}   & 0.39 & 17.81 & 0.46 & 0.55 & \underline{0.27} & 17.63 & 0.44 & 0.56 & \(\times\) & \(\times\) & \(\times\) & \(\times\) & \(\times\) & \(\times\) & \(\times\) & \(\times\) \\
AnySplat~\cite{anysplat}    & 0.54 & \underline{20.05} & 0.56 & \underline{0.38} & 2.59 & \cellcolor{tabhl}\textbf{20.98} & 0.52 & \underline{0.38} & 5.78 & 24.24 & 0.41 & 0.40 & 2.97 & 21.76 & 0.50 & \underline{0.39} \\
\textbf{Ours} & 0.24 & \cellcolor{tabhl}\textbf{21.71} & 0.57 & \cellcolor{tabhl}\textbf{0.31} & 0.54 & \underline{20.41} & \underline{0.56} & \cellcolor{tabhl}\textbf{0.37} & 4.85 & 29.83 & 0.55 & 0.41 & \underline{1.88} & \cellcolor{tabhl}\textbf{23.98} & 0.56 & \cellcolor{tabhl}\textbf{0.36} \\ \hline
\end{tabular}
}
\end{table*}

\section{Experiments}

\subsection{Implementation Details}
\label{sec:implementation}

NemoSplat is implemented in PyTorch and use the gsplat~\cite{gsplat} CUDA Library. We employ pre-trained DINOv2-ViT-L14~\cite{dinov2} encoder alongside a 24-layer causal alternating attention transformer~\cite{streamvggt}. To establish robust initial priors, geometry estimator and Gaussian predictor are loaded with pre-trained weights from StreamVGGT~\cite{streamvggt} and AnySplat~\cite{anysplat}, respectively. All other heads are randomly initialized. In total, the full model is approximately 1.22 Billion parameters. 

To achieve stable disentanglement, we employ a progressive two-stage training strategy. Stage 1 optimizes the Promptable Dynamic Disentangler to embed initial semantic and boundary priors. Stage 2 synergistically optimizes the full pipeline for structural geometry and media-aware rendering, where the visual backbone is frozen and LoRA~\cite{lora} is applied to maintain geometric stability. Our model is trained on 5 NVIDIA RTX 4090D GPUs, taking approximately 1 day for Stage 1 and 2 days for Stage 2. Detailed training configurations, and hyperparameters are provided in the supplementary material.


\textbf{Datasets.} We construct a large-scale underwater dataset with 256 training sequences (155K frames) and 20 evaluation scenes (partially illustrated in Fig.~\ref{fig:eval-set}). Specifically, the training corpus spans 9 diverse geographic domains and employs a 13-class taxonomy to systematically capture complex marine life behaviors. Training data includes depth, water, and dynamic masks annotated via a human-refined SAM3~\cite{sam3} and SAHI~\cite{sahi} pipeline to ensure robust extraction of small or distant objects. For evaluation, 6 synthetic UE5 sequences provide precise ground truth. Further details are in the supplementary material.

\begin{table*}[t]
\centering
\caption{Quantitative evaluation of novel view synthesis on 14 real-world aquatic sequences. We report PSNR, SSIM, and LPIPS metrics across all scenes alongside their overall averages. The best results are \colorbox{tabhl}{\textbf{bold}}, and the second best are \underline{underlined}.}
\label{tab:exp_realworld}
\renewcommand\arraystretch{1.2}
\resizebox{0.90\textwidth}{!}{
\begin{tabular}{llccccccccccccccc}
\toprule
\textbf{Method} & \textbf{Metrics}  & \textbf{01} & \textbf{02} & \textbf{03} & \textbf{04} & \textbf{05} & \textbf{06} & \textbf{07} & \textbf{08} & \textbf{09} & \textbf{10} & \textbf{11} & \textbf{12} & \textbf{13} & \textbf{14} & \textbf{Avg.} \\ 
\cmidrule(r){1-1} \cmidrule(r){2-2} \cmidrule(lr){3-16} \cmidrule(l){17-17}
\multirow{3}{*}{WildGS-SLAM~\cite{WildGS-SLAM}} 
    & PSNR$\uparrow$   & 19.31 & 18.13 & \cellcolor{tabhl}\textbf{20.85} & \underline{19.92} & \underline{19.24} & 21.75 & 19.79 & 16.08 & 21.40 & 19.38 & \cellcolor{tabhl}\textbf{20.79} & \cellcolor{tabhl}\textbf{21.64} & \underline{16.29} & 13.50 & 19.15 \\
    & SSIM$\uparrow$    & 0.56 & 0.65 & \cellcolor{tabhl}\textbf{0.74} & 0.62 & 0.52 & 0.74 & 0.62 & 0.42 & \underline{0.68} & 0.49 & \cellcolor{tabhl}\textbf{0.69} & \cellcolor{tabhl}\textbf{0.78} & \underline{0.58} & 0.37 & \underline{0.60}  \\
    & LPIPS$\downarrow$  & 0.50 & 0.52 & \underline{0.38} & \underline{0.37} & \underline{0.44} & 0.42 & 0.44 & 0.50 & \underline{0.35} & \underline{0.36} & \cellcolor{tabhl}\textbf{0.28} & \underline{0.25} & \underline{0.41} & 0.58 & \underline{0.41}
     \\ \hline
\multirow{3}{*}{Droid-W~\cite{DROIDW}} 
    & PSNR$\uparrow$ & \underline{20.59} & \underline{19.83} & \underline{18.61} & 18.73 & 17.77 & 20.50 & 17.05 & \underline{16.71} & 18.43 & 19.05 & 17.90 & 21.05 & 13.20 & 14.32 & 18.12 \\
    & SSIM$\uparrow$ & \cellcolor{tabhl}\textbf{0.73} & \cellcolor{tabhl}\textbf{0.69} & \underline{0.51} & 0.55 & 0.40 & 0.68 & 0.49 & \underline{0.50} & 0.60 & 0.46 & \underline{0.63} & \underline{0.72} & 0.50 & \underline{0.43} & 0.56  \\
    & LPIPS$\downarrow$  & 0.37 & \underline{0.39} & 0.46 & \underline{0.37} & 0.52 & 0.51 & 0.58 & \underline{0.42} & 0.43 & 0.39 & 0.36 & 0.27 & 0.50 & 0.57 & 0.44 \\ \hline
\multirow{3}{*}{YoNoSplat~\cite{yonosplat}} 
    & PSNR$\uparrow$  & 15.39 & 16.73 & 15.44 & 16.71 & 14.83 & 20.06 & 16.43 & 13.05 & 18.17 & 16.58 & 12.22 & 15.05 & 13.75 & 14.52 & 15.64 \\
    & SSIM$\uparrow$  & 0.44 & 0.35 & 0.33 & 0.42 & 0.33 & 0.66 & 0.54 & 0.38 & 0.52 & 0.37 & 0.35 & 0.41 & 0.34 & 0.35 & 0.41  \\
    & LPIPS$\downarrow$  & 0.67 & 0.64 & 0.63 & 0.62 & 0.69 & 0.44 & 0.56 & 0.58 & 0.53 & 0.56 & 0.64 & 0.57 & 0.60 & 0.62 & 0.60 \\ \hline
\multirow{3}{*}{AnySplat~\cite{anysplat}} 
    & PSNR$\uparrow$   & 20.43 & 18.28 & 17.69 & 19.80 & 18.83 & \underline{22.04} & \underline{20.19} & 16.44 & \underline{21.54} & \underline{19.84} & 20.45 & 20.86 & 15.14 & \underline{14.80} & \underline{19.22} \\
    & SSIM$\uparrow$ & \underline{0.68} & 0.47 & 0.46 & \underline{0.63} & \underline{0.55} & \underline{0.75} & \cellcolor{tabhl}\textbf{0.72} & 0.48 & \cellcolor{tabhl}\textbf{0.73} & \underline{0.55} & \cellcolor{tabhl}\textbf{0.69} & 0.68 & 0.46 & 0.42 & \underline{0.60}  \\
    & LPIPS$\downarrow$  & \underline{0.35} & 0.50 & 0.47 & 0.48 & 0.49 & \underline{0.40} & \underline{0.39} & 0.45 & \underline{0.35} & 0.40 & 0.32 & 0.30 & 0.49 & \underline{0.56} & 0.42  \\ \hline
\multirow{3}{*}{\textbf{Ours}} 
    & PSNR$\uparrow$   & \cellcolor{tabhl}\textbf{20.98} & \cellcolor{tabhl}\textbf{22.35} & 18.21 & \cellcolor{tabhl}\textbf{21.67} & \cellcolor{tabhl}\textbf{20.10} & \cellcolor{tabhl}\textbf{22.79} & \cellcolor{tabhl}\textbf{22.89} & \cellcolor{tabhl}\textbf{19.99} & \cellcolor{tabhl}\textbf{24.84} & \cellcolor{tabhl}\textbf{21.45} & \underline{20.72} & \underline{21.53} & \cellcolor{tabhl}\textbf{21.21} & \cellcolor{tabhl}\textbf{22.13} & \cellcolor{tabhl}\textbf{21.58} \\
    & SSIM$\uparrow$    & 0.60 & \cellcolor{tabhl}\textbf{0.69} & 0.48 & \cellcolor{tabhl}\textbf{0.68} & \cellcolor{tabhl}\textbf{0.57} & \cellcolor{tabhl}\textbf{0.82} & \underline{0.67} & \cellcolor{tabhl}\textbf{0.75} & \cellcolor{tabhl}\textbf{0.73} & \cellcolor{tabhl}\textbf{0.63} & \cellcolor{tabhl}\textbf{0.69} & 0.61 & \cellcolor{tabhl}\textbf{0.70} & \cellcolor{tabhl}\textbf{0.75} & \cellcolor{tabhl}\textbf{0.68}  \\
    & LPIPS$\downarrow$ & \cellcolor{tabhl}\textbf{0.31} & \cellcolor{tabhl}\textbf{0.23} & \cellcolor{tabhl}\textbf{0.36} & \cellcolor{tabhl}\textbf{0.24} & \cellcolor{tabhl}\textbf{0.35} & \cellcolor{tabhl}\textbf{0.25} & \cellcolor{tabhl}\textbf{0.30} & \cellcolor{tabhl}\textbf{0.18} & \cellcolor{tabhl}\textbf{0.23} & \cellcolor{tabhl}\textbf{0.27} & \underline{0.29} & \cellcolor{tabhl}\textbf{0.24} & \cellcolor{tabhl}\textbf{0.16} & \cellcolor{tabhl}\textbf{0.21} & \cellcolor{tabhl}\textbf{0.26}  \\ \bottomrule
\end{tabular}
}
\end{table*}

\subsection{Experimental Setup}
\label{sec:setup}

\textbf{Metrics.} To comprehensively evaluate our proposed NemoSplat, we conduct experiments on our evaluation dataset. First, we benchmark the tracking and render performance on 6 challenging synthetic sequences. Pose accuracy is evaluated using Absolute Trajectory Error (ATE RMSE, $m$), while rendering quality is assessed using PSNR, SSIM~\cite{ssim}, and LPIPS~\cite{lpips}. Second, to demonstrate our generalizability in real-world marine environment, we evaluate the novel view synthesis quality on 14 diverse real-world underwater sequences.

\textbf{Baselines.} We compare NemoSplat against six state-of-the-art baselines across various novel view synthesis and reconstruction paradigms. Specifically, we evaluate two feed-forward visual foundation models, VGGT~\cite{vggt} and StreamVGGT~\cite{streamvggt}, which estimate geometry and poses. We also benchmark against cutting-edge feed-forward 3DGS frameworks, YonoSplat~\cite{yonosplat} and AnySplat~\cite{anysplat}, which enable rapid reconstruction without per-scene optimization. Lastly, we compare with optimization-based dynamic Gaussian-based SLAM systems, Droid-W~\cite{DROIDW} and WildGS-SLAM~\cite{WildGS-SLAM}. For fair comparison, we provide the SLAM baselines with VGGT-estimated camera intrinsics~\cite{vggt}. All evaluations run on a single RTX 4090D GPU.

\subsection{Evaluation of Camera Tracking}
\label{sec:evaluate_tracking}

As detailed in Tab.~\ref{tab:exp_synthetic}, we quantitatively benchmark the camera tracking performance of NemoSplat against six competitive baselines on our synthetic set, which comprises the BoulderShore, Coral, and Deepsea synthetic environments, with two distinct sequences per scene. Our method extracts stable camera trajectories in challenging dynamic environments despite uncalibrated inputs. NemoSplat achieves a highly competitive average ATE of $1.88$ m, consistently outperforming other feed-forward pipelines like VGGT ($2.12$ m) and StreamVGGT ($2.06$ m). This superior performance is driven by the joint optimization of the geometry estimator and the media-aware rendering pipeline, which enhances resilience to transient moving distractors in dynamic underwater environments. Furthermore, we observe a distinct dichotomy in tracking behaviors depending on scene characteristics. In the ``Coral'' sequences, which feature large inter-frame motion and broad fields of view, the optical flow estimation essential to DROID-W and WildGS-SLAM struggles to establish stable pixel associations, resulting in high trajectory drift. Feed-forward models, conversely, leverage diverse learned spatial priors to handle such drastic viewpoint changes effectively. On the other hand, in the light-deprived ``Deepsea'' sequences, the pre-trained image encoders of feed-forward architectures fail to extract reliable semantic features. Under these extreme low-visibility conditions, optical-flow based SLAM methods successfully capitalize on raw inter-frame pixel variations to maintain robust tracking.

\subsection{Evaluation of Novel View Synthesis}
\label{sec:evaluate_nvs}

We comprehensively evaluate the novel view synthesis quality across both simulated and real-world environments, as shown in Tab.~\ref{tab:exp_synthetic} and Tab.~\ref{tab:exp_realworld}. On the synthetic dataset, our feed-forward model achieves the best visual quality, yielding the highest overall PSNR of $23.98$ dB and the lowest LPIPS of $0.36$. These results consistently outperform the runner-up methods, taking a $0.47$ dB lead in PSNR over DROID-W and reducing LPIPS by $0.03$ compared to AnySplat. 

\begin{figure*}[t]
  \centering
  \includegraphics[width=0.95\linewidth]{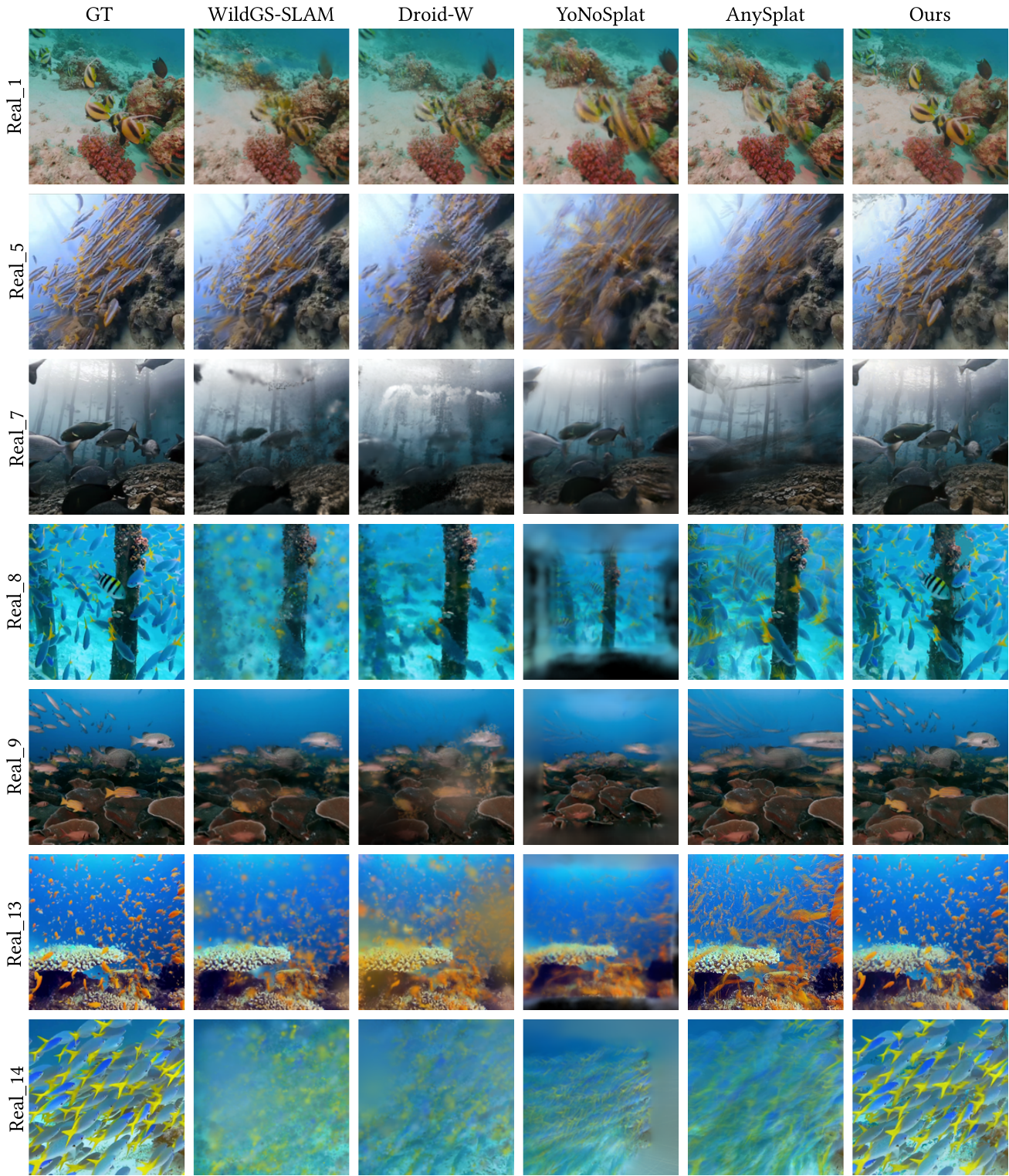}
   \Description{Figures showing rendering results of baseline models and NemoSplat.}
    \caption{Qualitative comparison of novel view synthesis on highly degraded real-world aquatic sequences. While conventional feed-forward modules and SLAM-based baselines suffer from severe ghosting, topological blurring, and visual artifacts caused by unconstrained dynamic entities (e.g., schools of swimming fish), NemoSplat explicitly decouples transient motion from the static topology. This disentanglement enables the synthesis of crisp, temporally consistent, and artifact-free novel views.}
  \label{fig:compare_realworld}
\end{figure*}

\begin{figure*}[t]
  \centering
  \includegraphics[width=0.95\linewidth]{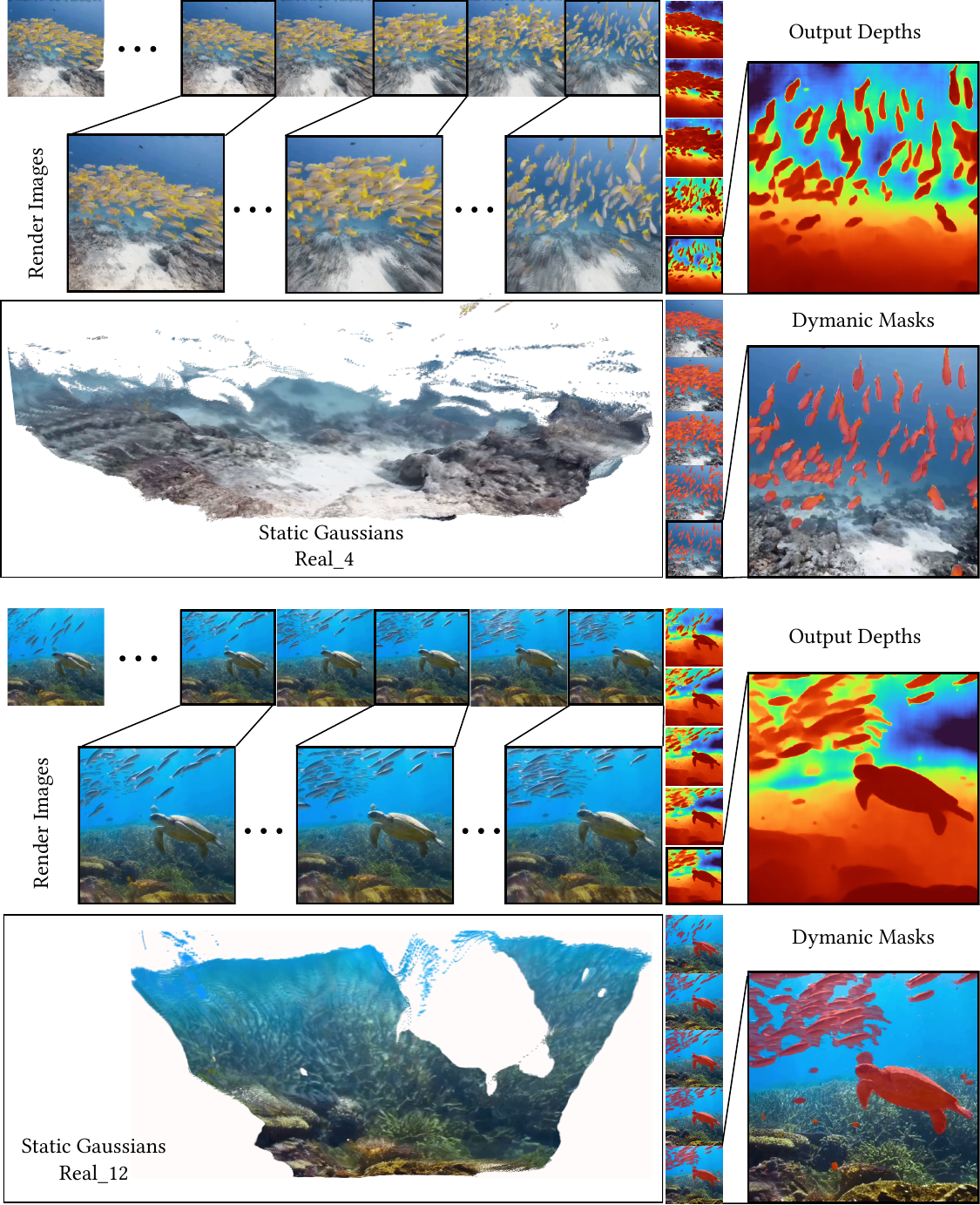}
    \caption{Reconstruction results on real-world aquatic sequences. Our method disentangles static Gaussian models from dynamic underwater scenes, providing comprehensive outputs including RGB rendering images, depth maps and dynamic masks.}
  \label{fig:viz_realworld}
\end{figure*}

Moving to the significantly more challenging real-world scene, comprising 14 highly degraded aquatic sequences, NemoSplat comprehensively surpasses all baselines across all metrics. Specifically, it establishes state-of-the-art averages in PSNR ($21.58$ dB), SSIM ($0.68$), and LPIPS ($0.26$). Our method achieves the highest PSNR and SSIM, exceeding the second-best AnySplat ($19.22$ dB, 0.60) by $2.36$ dB and 0.08. Most impressively, our LPIPS is substantially reduced, achieving a $36.6\%$ relative error reduction compared to WildGS-SLAM ($0.41$). This pronounced performance gap is highly attributable to the complex nature of real-world scenes, containing abundant dynamic entities such as massive swimming fish schools. In conventional reconstruction pipelines, these unconstrained moving objects inevitably introduce severe rendering artifacts, topological blurring, and ghosting effects. As shown in Fig.~\ref{fig:compare_realworld}, competing methods exhibit severe artifacts and blurring, whereas NemoSplat cleanly separates dynamic and static components to produce clear and accurate renderings.
NemoSplat elegantly overcomes this bottleneck by explicitly extracting precise dynamic masks, incorporating text prompts as an optional guidance. By effectively decoupling per-frame dynamic Gaussian primitives from the static scene topology, our approach eliminates the risk of baking transient motion into the global background, thereby realizing highly accurate and temporally consistent rendering of complex time-varying scenes. Fig.~\ref{fig:viz_realworld} illustrates the rich, high-fidelity reconstruction results achieved by our method across real-world marine environments.

\subsection{Evaluation of Descattering}
\label{sec:evaluate_descattering}

\begin{figure}[t]
    \centering
    \includegraphics[width=0.95\linewidth]{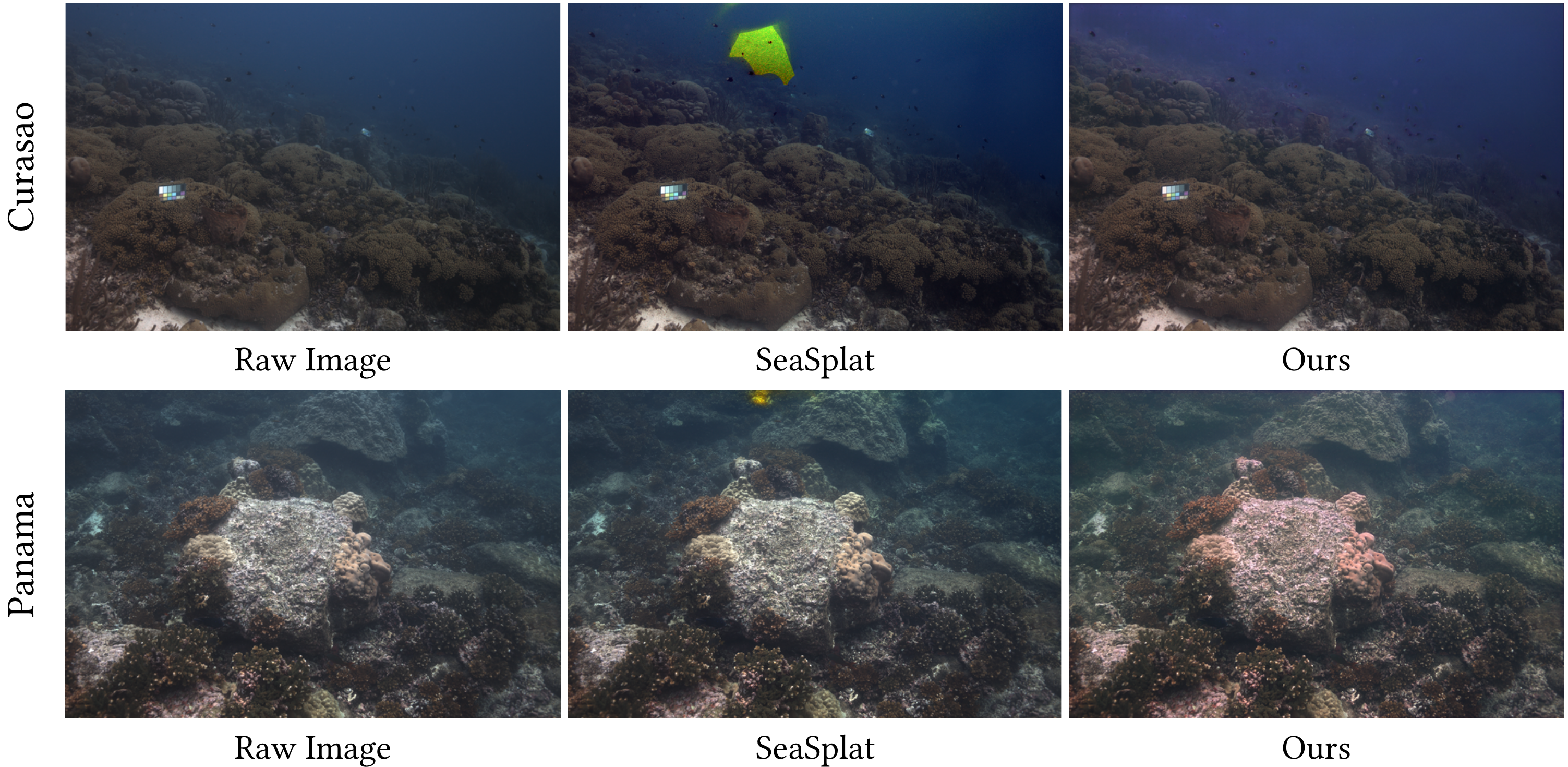}
    \Description{Figures showing restored images on SeaThru-NeRF dataset.}
    \caption{Restored images on SeaThru-NeRF~\cite{seathru-nerf} dataset. Our approach recovers more natural colors while preserving clearer details in distant regions.}
    \label{fig:descattering_qualitative}
\end{figure}

To validate the effectiveness in predicting physical water medium parameters, we evaluate our descattering capabilities on the SeaThru-NeRF~\cite{seathru-nerf} dataset, shown in Fig.~\ref{fig:descattering_qualitative}. By a single forward pass, NemoSplat simultaneously infers the complete set of water medium parameters, including a per-frame global veiling light $\boldsymbol{B}^{\infty}$ and per-pixel estimations for direct signal attenuation $\boldsymbol{\beta}^D$ and volumetric backscatter $\boldsymbol{\beta}^B$. We compare our approach against the state-of-the-art method SeaSplat~\cite{seasplat}. SeaSplat was run for 1k optimization iterations, taking approximately 6 minutes. 
In contrast, our approach requires fewer than 10 seconds to recover more reliable colors than SeaSplat, with clearer distant structures. Although per-scene optimization-based 3DGS methods like SeaSplat can achieve superior descattering results given extended iterations, our media-aware Gaussian predictor infers highly reasonable water medium parameters in a single forward pass. This demonstrates the immense potential of our method, providing a novel and rapid paradigm for mitigating underwater attenuation and scattering.

\section{Conclusion}
\label{sec:conclusion}

We presented NemoSplat, a robust feed-forward 4D Gaussian Splatting framework tailored for uncalibrated, dynamically complex, and severely degraded aquatic environments. By coupling a media-aware rendering pipeline with optional text-guided semantic reasoning, our method elegantly removes dense underwater scattering and decouples transient entities from the static geometry. Experiments on our dataset show that NemoSplat achieve state-of-the-art performance on both artifact-free novel view synthesis and robust camera tracking. Despite these strong capabilities, integrating heavy semantic masking and physical media modules causes substantial GPU memory consumption during training, remaining a primary limitation. Future efforts will focus on optimizing architectural memory efficiency to unlock broader real-time deployments in autonomous underwater vehicle (AUV) navigation and marine exploration.

\bibliographystyle{ACM-Reference-Format}
\bibliography{reference}

\newpage

\newpage

\end{document}